\documentclass{article}

\PassOptionsToPackage{numbers, square}{natbib}
\usepackage[preprint]{neurips_2024}
\usepackage[whole]{bxcjkjatype} % add for Japanese use

\usepackage[utf8]{inputenc} % allow utf-8 input
\usepackage[T1]{fontenc}    % use 8-bit T1 fonts
\usepackage[hidelinks]{hyperref}
\usepackage{url}            % simple URL typesetting
\usepackage{booktabs}       % professional-quality tables
\usepackage{amsfonts}       % blackboard math symbols
\usepackage{nicefrac}       % compact symbols for 1/2, etc.
\usepackage{microtype}      % microtypography
\usepackage{xcolor}         % colors

\usepackage{multirow}
\usepackage{graphicx}
\usepackage{bm}
\usepackage{url}
\usepackage{pifont}
\usepackage{mathtools}
\usepackage{adjustbox}
\usepackage{colortbl}
\usepackage{array}
\usepackage{textcomp}

\usepackage{algpseudocode}
\usepackage{algorithm}

\newcommand\eg{\textit{e.g.}}
\newcommand\ie{\textit{i.e.}}
\newcommand\etal{\textit{et al.}}
\newcommand\cf{\textit{cf.}}

\hypersetup{colorlinks,linkcolor={red},citecolor={blue}}

\title{Language-guided Detection and Mitigation of Unknown Dataset Bias}
\author{
    Zaiying Zhao$^1$, Soichiro Kumano$^1$, Toshihiko Yamasaki$^1$\\
    $^1$The University of Tokyo\\
    {\tt \small \{zhao,kumano,yamasaki\}@cvm.t.u-tokyo.ac.jp}\\ 
}

\begin{document}
\maketitle

\begin{abstract}
  Dataset bias is a significant problem in training fair classifiers. 
  When attributes unrelated to classification exhibit strong biases
  % (i.e., spurious correlation) 
  towards certain classes, classifiers trained on such dataset may overfit to these bias attributes, substantially reducing the accuracy for minority groups. 
  Mitigation techniques can be categorized according to the availability of bias information (\ie, prior knowledge). 
  % and the label of the bias attribute
  Although scenarios with unknown biases are better suited for real-world settings, previous work in this field often suffers from a lack of interpretability regarding biases and lower performance.
  %In addition, there exists a strong demand to ensure classifiers are free from any biases, making the autonomous detection of potential biases essential.
  In this study, we propose a framework to identify potential biases as keywords without prior knowledge based on the partial occurrence in the captions. 
  % and train debiased classifier using the acquired bias information. In particular, 
  We further propose two debiasing methods: (a) handing over to an existing debiasing approach which requires prior knowledge by assigning pseudo-labels, and (b) employing data augmentation via text-to-image generative models, using acquired bias keywords as prompts. 
  % Our framework is well-suited for real-world applications by linguistically presenting biases, and
  Despite its simplicity, experimental results show that our framework 
 not only outperforms existing methods without prior knowledge, but also is even comparable with a method that assumes prior knowledge.
\end{abstract}

\section{Introduction}
Deep neural networks (DNNs) achieve remarkable performance in many fields and are widely applied in society when models are trained on carefully curated and balanced datasets.
On the other hand, models trained on biased datasets exhibit poor generalization performance, significantly compromising accuracy for minority groups. 
% For example, consider classifying horses versus camels, and the majority of horse images have grassland backgrounds while the majority of camel images have desert backgrounds.
%Training a classifier on such biased datasets allows it to easily minimize loss by learning predominant attributes in each class, and classifier inadvertently overfits to undesirable features like backgrounds.
%Despite achieving high average accuracy, such models severely degrade accuracy for minority groups (e.g., misclassifying a camel on grassland as horse). 
Specifically, it is well-known that object classifiers often focus significantly on background scenes. 
When recognizing ships, for example, water in the background serves as a strong clue for DNNs to identify the object as ships, often ending up with misrecognition on ships at land~\citep{Lapuschkin_2019}.
As another example, in facial image datasets, certain attributes are predominantly observed only in specific genders or age groups (\eg, blonde hair is more common in females)~\citep{wang2019balanced}. 
These biases hinder recognition accuracy for minority groups.
Early attempts to mitigate biases,~\eg, data resampling~\citep{ramaswamy2021fair, li2019repair, byrd2019effect, qraitem2023bias, wang2020fairness, byrd2019effect, 10.5555/1293951.1293954} and explicit supervision~\citep{wang2019balanced, Sagawa*2020Distributionally, ross2017right, alvi2018turning, Tartaglione_2021, kim2019learning, NEURIPS2021_de8aa43e}, assume that prior knowledge of existing biases is available.
However, such an assumption is not well-suited in most real-world scenarios because it is difficult to identify biases within datasets in advance.

\begin{figure}[t]
  \centering
  \includegraphics[keepaspectratio, width=397pt]{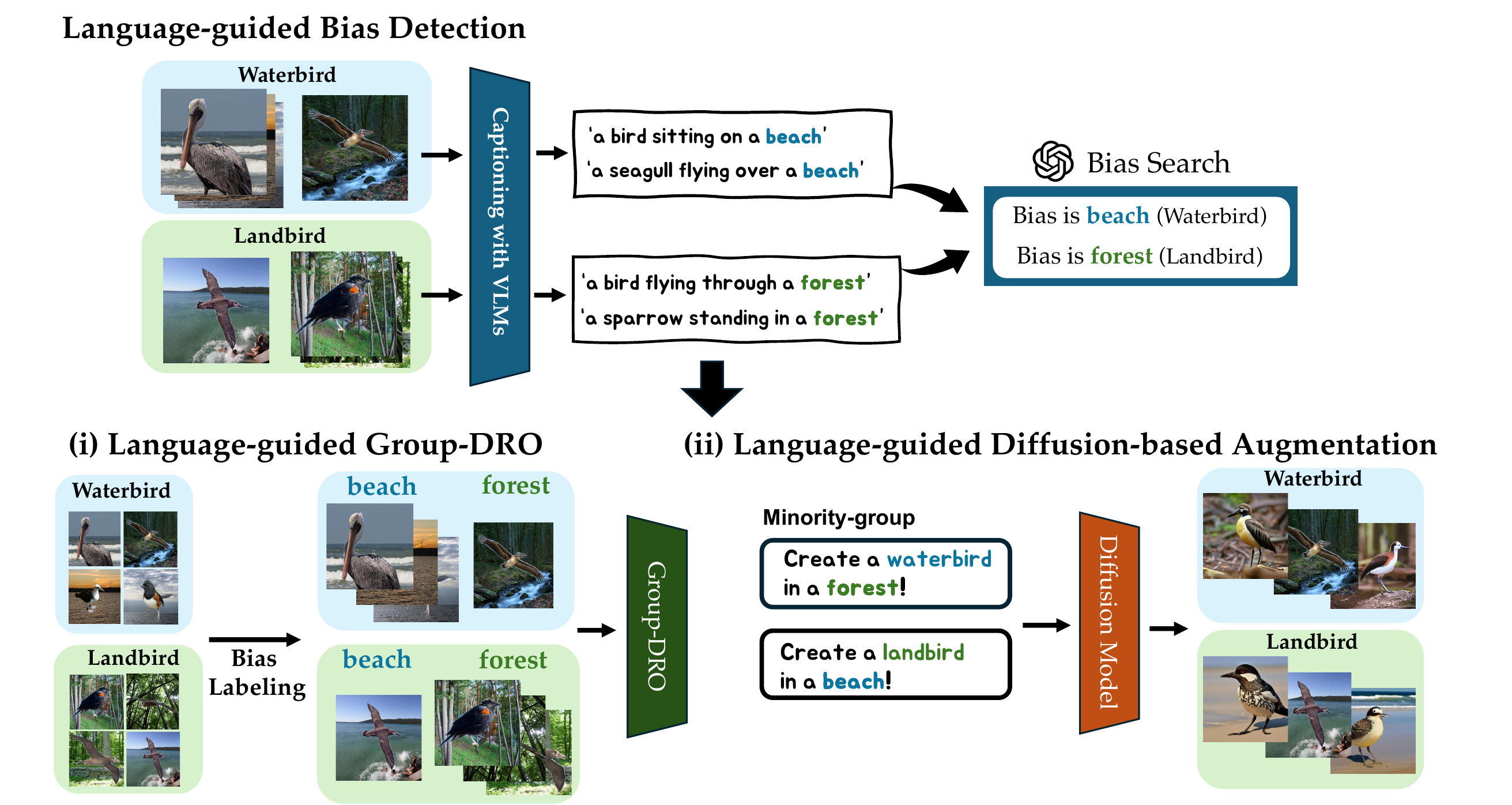}
  \vspace{-7mm}
  \caption{
  \small{\textbf{Overview of our framework.} First, we perform Language-guided Bias Detection, which identifies bias keywords using GPT-4 from captions generated by VLMs (\eg, BLIP). With these extracted keywords, we propose (i) Language-guided Group-DRO, which adapts Group-DRO by leveraging the extracted bias keywords as pseudo-bias labels, and (ii) Language-guided Diffusion-based Augmentation, which generates images for the minority groups leveraging the bias keywords as input prompts.
}
  }
  \label{fig:overview}
\end{figure}

To tackle unknown biases, recent work aims to detect biases through mispredicted samples~\citep{nam2020learning, liu2021just} or image features~\citep{Seo_2022_CVPR, zhang2022correctncontrast} from biased models.
Nonetheless, these methods have not yet achieved mitigation performance comparable to those using prior knowledge.
Moreover, detected biases are in implicit forms, such as image features or images themselves, requiring additional human interpretation for discernment.
In response to the issue of interpretability, debiasing methods utilizing vision-language models (VLMs) have been proposed~\citep{9880269, kim2024discovering}.
GALS~\citep{9880269} uses VLMs to project image regions corresponding to classes, and introduces a loss function based on the distance between image regions a model focuses on (\ie, attention map~\citep{dosovitskiy2021image}) and the projected regions. 
However, GALS has limited applicability when bias attributes cannot be visually separated from the target class (\eg, facial expressions or makeup in gender classification tasks) or when bias attributes cannot be represented in image regions, such as color. 
Notably, B2T~\citep{kim2024discovering} utilizes VLMs for image captioning and detects biases within a model as keywords from the captions of misclassified images. 
Yet, B2T employs conventional algorithms~\citep{CAMPOS2020257, 10.1007/978-3-319-76941-7_63, 10.1007/978-3-319-76941-7_80} for keyword extraction, leaving open the potential for even more accurate extraction with the use of large language models (LLMs).
Furthermore, in B2T, the extracted biases are not utilized in debiasing methods in the form of keywords.

In this study, inspired by GALS~\citep{9880269} and B2T~\citep{kim2024discovering}, we propose a framework that leverages VLMs and LLMs to identify potential bias keywords by detecting their partial emergence in image captions. Our main concepts entail that biases within images including visually indivisible attributes (\eg, color) are reflected in the captions generated by VLMs, addressing the limitations of GALS.
Additionally, the use of LLMs facilitates more accurate bias detection by identifying predominant keywords than B2T.
% Additionally, acquired bias keywords can inform the use of approaches applicable solely in settings with known biases and labeled attributes. Based on this concept, 
We further propose two powerful debiasing approaches: (i) \textbf{Language-guided Group-DRO} and (ii) \textbf{Language-guided Diffusion-based Augmentation}. (i) In Language-guided Group-DRO, we first assign pseudo-labels of the extracted biases to the dataset, replicating a scenario where prior knowledge is available. We then adapt the high-performance supervised learning method Group-DRO~\citep{Sagawa*2020Distributionally} with bias pseudo-labels. (ii) In Language-guided Diffusion-based Augmentation, we first construct prompts describing minority groups by utilizing extracted keywords. Using the prompts as input, image generation through Stable Diffusion~\citep{nokey} is then performed to increase the data for minority groups, mitigating dataset bias itself.
Our work is highly motivated by B2T~\citep{kim2024discovering} to identify visual biases using generated captions. However, unlike their work, we propose a more accurate keyword extraction method using GPT-4~\citep{openai2024gpt4} and also introduce a novel automatic diffusion-based augmentation scheme leveraging extracted biases in the form of keywords. 

We evaluated our framework on various biased benchmarks including CMNIST~\citep{arjovsky2020invariant}, Waterbirds~\citep{Sagawa*2020Distributionally} and CelebA~\citep{liu2015faceattributes}.
Despite the simplicity of our concepts, experimental results show that our debiasing methods outperform existing methods in the same settings where bias is unknown, and is even competitive with the method that requires prior knowledge.

Our contributions can be summarized as follows:
\begin{itemize}
    \item We propose a framework to extract potential dataset biases as keywords from captions using GPT-4, enabling more accurate and interpretable dataset bias detection in real-world settings.
    \item Leveraging the extracted bias keywords, we introduce two debiasing methods based on supervised learning and Diffusion-based augmentation.
    \item Our debiasing methods have outperformed existing methods without prior knowledge, and even achieved competitive performance with the method using prior knowledge.
\end{itemize}

% The rest of the paper follows this structure: Section~\ref{sec:related_work} reviews existing debiasing methods and explores the potential of VLMs for achieving more interpretable and high-performance bias mitigation. 
% Our framework is detailed in Section~\ref{sec:methodology}, and we conduct experiments and present the effectiveness of our framework in Section~\ref{sec:experiment}. 
% Finally, we conclude our paper in Section~\ref{sec:conclusion}. 
The code will be made publicly available upon acceptance of the paper.

\section{Related work}
\label{sec:related_work}
\textbf{Debiasing known biases.}
Some existing methods assume that the labels of biased attributes are provided as prior knowledge. 
Existing work in this field can be divided into two approaches. The first approach aims to directly address dataset biases by eliminating them~\citep{wang2020fairness, byrd2019effect, 10.5555/1293951.1293954, ramaswamy2021fair, li2019repair, qraitem2023bias}, while the second approach aims to prevent learning biases~\citep{ross2017right, alvi2018turning, Tartaglione_2021, kim2019learning, wang2019balanced, NEURIPS2021_de8aa43e, Sagawa*2020Distributionally}.
Ramaswamy~\etal~\citep{ramaswamy2021fair} apply data augmentation by using Generative Adversarial Networks (GANs)~\cite{goodfellow2014generative} to construct balanced datasets. 
Li~\etal~\citep{li2019repair} propose data resampling method adjusting the resampling weights using stochastic gradient descent (SGD) algorithm during the training process, and Qraitem~\etal~\citep{qraitem2023bias} propose mitigation of multiple biases by subsampling datasets to emulate the bias distributions of each class.
In terms of training unbiased classifiers on biased data, Tartaglione~\etal~\citep{Tartaglione_2021} define two regularization terms: one to eliminate correlations among data sharing the same bias labels and the other to strengthen correlations among data with different bias labels.
% incorporating these terms into loss function by weighting appropriately.
Kim~\etal~\citep{kim2019learning} introduce a regularization loss grounded in mutual information, encouraging feature embedding networks to eliminate bias information.
Wang~\etal~\citep{wang2019balanced} adapt an adversarial approach to visually remove bias attribute from images, and Hong~\etal~\citep{NEURIPS2021_de8aa43e} introduce two losses for debiasing: one based on a constrastive learning framework to leverage bias labels, and the other based on regression balancing the distribution of bias.
In particular, Sagawa~\etal~\citep{Sagawa*2020Distributionally} extend the idea of distributionally robust optimization (DRO)~\cite{shafieezadehabadeh2015distributionally, hashimoto2018fairness} that considers worst-case distributions, and propose Group-DRO by applying it to the worst-group within the dataset, achieving state-of-the-art generalization performance.
These methods hinge on prior knowledge and assume the existence of dataset bias. 
However, in practice, detecting biases requires thorough dataset analysis, making this assumption challenging. 
Moreover, the exploitation of annotations for bias attributes renders these methods impractical for real-world applications.

\textbf{Debiasing unknown biases.}
Debiasing unknown biases is a far more challenging problem compared to situations where prior knowledge is available. 
% In this context, many existing methods focus on addressing biases present in image features or on learning unbiased classifiers from features, including the weights derived from intentionally trained biased classifiers.
In this context, many existing methods aim to tackle biases found in image features or focus on developing unbiased classifiers using features obtained from intentionally trained biased classifiers.
In LfF~\citep{nam2020learning}, two models are simultaneously trained, with one intentionally overfitting, and the loss gradients of that model are used to reweight the losses of the other model. 
Similarly, Liu~\etal~\citep{liu2021just} introduce a two-step method called JTT, which initially trains a model with biases and then follows up with a second model that prioritizes examples misclassified by the initial model. 
BPA~\citep{Seo_2022_CVPR} estimates bias pseudo-attributes by clustering in the feature space and performs reweighting based on size and loss of each cluster. 
Zhang~\etal~\citep{zhang2022correctncontrast} propose CNC, which 
% sets reference points in the feature space and 
performs contrastive learning to differentiate feature points with the same predicted labels while bringing feature points with different predicted labels closer together.
Nevertheless, these approaches lack interpretability concerning biases and often result in lower accuracy compared to debiasing methods used in scenarios where prior knowledge is available.

\textbf{Leveraging VLMs.}
In recent years, VLMs have made significant advances and are being applied across various fields of research~\citep{zhang2024visionlanguage, hu2022make, zhou2022learning, changpinyo2022need, miyai2024locoop, Avrahami_2022, NEURIPS2022_5bf2b802}. 
In terms of training unbiased classifiers, Petryk \etal~\citep{9880269} propose GALS, a debiasing method utilizing VLMs to project image regions corresponding to classes and achieve bias mitigation with a certain level of interpretability by training the model to focus on the projected image regions (\ie, attention map~\citep{dosovitskiy2021image}).
However, GALS is less effective when bias attributes are visually indivisible from the target class (\eg, facial expressions or makeup in gender classification tasks) or when bias attributes cannot be represented in image regions, such as color. 
% Additionally, attention maps do not indicate the bias itself, and therefore do not address the fundamental problem.
Notably, Kim \etal~\citep{kim2024discovering} propose B2T, a language-guided framework that detects the biases to which models overfit by analyzing the captions of misclassified images. 
While B2T achieves bias detection with high interpretability through keyword extraction, it relies on conventional algorithms~\citep{CAMPOS2020257, 10.1007/978-3-319-76941-7_63, 10.1007/978-3-319-76941-7_80}, leaving room for potentially more accurate extraction through the use of LLMs.

Drawing from the findings of B2T~\citep{kim2024discovering} regarding the utility of captions in detecting bias, we introduce a framework named language-guided bias detection and mitigation.
% as shown in Figure~\ref{fig:overview}.
Our approach is based on a simple intuition that, in contrast to GALS~\citep{9880269} which projects text onto images, utilizing VLMs in an image captioning approach can overcome the limitations present in GALS, while at the same time addressing the poor interpretability of existing methods debiasing unknown bias.
In this framework, we first extract potential biases present in the dataset as keywords from the captions using GPT-4~\citep{openai2024gpt4}. Unlike B2T, which uses conventional approaches for keyword extraction, our use of GPT-4 offers the potential for more accurate identification of biases.
Within this detection framework, classes and visually indivisible attributes can be linguistically separated within captions, thus overcoming the first limitation of GALS. 
Furthermore, attributes that cannot be fully represented in image regions, such as color, can be reflected in the captions, thereby addressing the second limitation as well.
Utilizing the extracted bias keywords, this framework subsequently provides two independent debiasing methods.
In the first debiasing method, we perform pseudo-annotation of detected biases to establish a setting where biases are known. This allows us to utilize Group-DRO~\citep{Sagawa*2020Distributionally}, a high-performance debiasing method that requires prior knowledge.
In the second debiasing method, we aim to make the most of the characteristic that biases are identified as keywords by utilizing them as prompts to conduct generative model-based data augmentation.

\section{Method}
\label{sec:methodology}
In this section, we describe our methods for extracting and mitigating dataset biases. First, in Section~\ref{sec:3.1}, we define the problem of dataset biases. Then, we describe language-guided bias detection using VLMs in Section~\ref{sec:3.2}, and two language-guided debiasing methods utilizing the detected bias keywords in Section~\ref{sec:3.3}. Overview of our framework is shown in Figure~\ref{fig:overview}.

\subsection{Problem definition}
\label{sec:3.1}
Let $m\in\mathbb{N}$ attributes within dataset images be $a_1, \dots ,a_m$ and the set of attributes be $\mathcal{A}\coloneqq\{a_1, \dots ,a_m\}$, where $a_t$ is the target attribute to be classified. 
In cases where there is a strong correlation (\ie, spurious correlation) between a particular attribute $a_b\in\mathcal{A}$ and $a_t$ (as indicated by the conditional entropy $H(a_t|a_b) \approx 0$), the classifier may predict $a_t$ by mistakenly learning $a_b$. 
This attribute $a_b$ is termed a bias attribute and is recognized as dataset bias.
In this study, we identify bias attribute $a_b$ and debias it either by refraining from learning $a_b$ or by mitigating the spurious correlation between $a_b$ and $a_t$.

\subsection{Language-guided bias detection}
\label{sec:3.2}
\textbf{Keyword extraction leveraging GPT-4.}
First, we utilize a pretrained VLM to generate captions for all training images in a dataset. 
If certain words appear disproportionately in the captions for specific classes, the model tends to overfit to these attributes during training. 
Therefore, by leveraging GPT-4~\citep{openai2024gpt4}, we extract candidates of biases for each class based on the partial occurence in the captions. 
Note that attempts to extract biases from captions have already been made with B2T~\citep{kim2024discovering}, and its effectiveness has also been demonstrated. 
However, while B2T uses conventional algorithms~\citep{CAMPOS2020257, 10.1007/978-3-319-76941-7_63, 10.1007/978-3-319-76941-7_80} for keyword extraction, our framework employs LLMs such as GPT-4, which have the potential to achieve more accurate bias identification.

\textbf{Validation of extracted keywords.}
Extracted bias keywords for each class do not necessarily correspond to potential bias attributes in the dataset. For example, when performing binary classification of gender in a facial image dataset, the word ``people'' is common across both gender classes.
% if the word ``people'' is extracted as a candidate of bias attribute for each class, it is not valid since it is common across both gender classes.
Therefore, we verify whether the extracted bias keywords for each class are specific to that class by using the text-image similarity of CLIP~\citep{radford2021trans}.
In multi-class classification with $C\in\mathbb{N}$ classes, let the set of classes be $\mathcal{C} \coloneqq \{1, \dots C\}$
and subset of data for class $c\in\mathcal{C}$ be $\chi_c \subset \mathbb{R}^{H\times W\times 3}$.
The text-image similarity $S_{\mathrm{CLIP}}:\mathbb{R}^d\times\mathbb{R}^{H\times W\times 3}\to\mathbb{R}$ between an extracted bias keyword $\bm{w}\in\mathbb{R}^d$ and a subset of data is defined as the average inner product of the normalized embeddings of CLIP text encoder $\bm{f}_\mathrm{text}:\mathbb{R}^d\to\mathbb{R}^O$ and CLIP image encoder $\bm{f}_\mathrm{image}:\mathbb{R}^{H\times W\times 3}\to\mathbb{R}^O$, as described below:
\begin{equation}
    S_{\mathrm{CLIP}}(\bm{w},\chi_c) = \frac{1}{|\chi_c|}\sum_{\bm{x} \in \chi_c}{\bm{f}_\mathrm{text}(\bm{w}) \cdot \bm{f}_\mathrm{image}(\bm{x})}.
    \label{eq:score_similarity} 
\end{equation}
In this context, the class-specificity score $S_{\mathrm{specific}}:\mathbb{R}^d\times\mathbb{R}^{H\times W\times 3}\to\mathbb{R}$ of the bias attributes extracted for each class is defined as Eq. (\ref{eq:score_specificity}). This class-specificity score $S_{\mathrm{specific}}$ attains a high value when the extracted bias attribute demonstrates a stronger similarity to the images of that class. In our proposed method, we determine the final bias attributes by selecting the candidates with high class-specificity scores $S_{\mathrm{specific}}$.
\begin{equation}
    S_{\mathrm{specific}}(\bm{w}, \chi_c) = S_{\mathrm{CLIP}}(\bm{w},\chi_c) - \frac{1}{C-1}\sum_{c^{\prime} \in \mathcal{C}\setminus\{c\}}S_{\mathrm{CLIP}}(\bm{w},\chi_{c^{\prime}}).
    \label{eq:score_specificity} 
\end{equation}

\subsection{Bias mitigation leveraging extracted bias keywords}
\label{sec:3.3}
The framework developed so far focuses on extracting potential biases in a dataset as keywords. 
Moving forward, we propose two bias mitigation methods using extracted bias keywords.

\textbf{Lg-DRO: Language-guided Group-DRO.}
Existing high-performance debiasing methods such as Group-DRO~\citep{Sagawa*2020Distributionally} rely on not only prior knowledge of biases but also per-sample bias annotations in a dataset. 
Therefore, to apply these methods to unknow bias problems, we assign pseudo-labels of the extracted bias attributes to the dataset. 
This pseudo-bias annotation is performed through CLIP Zero-Shot classification~\citep{radford2021trans}, computing the similarity between prompts and images.
The prompts are constructed for each class based on the class and the extracted bias attributes, for example, as ``a photo of a \{Class\} in \{Attribute\}'' (more examples of prompt templates are in Appendix~\ref{appendix:settings}).
The created prompts are grouped based on the class and bias attributes, and
% For example, in the case of binary classification between camels and horses, four groups are formed based on the class and background information.
zero-shot classification is then performed to determine which group each training image belongs to.
% この次にzero-shot classificationの数式を入れる？ → 恐らく既知として省いてよい、入れても自己満になりそう
This grouping simply involves pseudo-annotation of the training images according to the bias attributes they possess, enabling the use of existing high-performance debiasing methods based on the assigned labels. 
In this paper, we apply Group-DRO~\citep{Sagawa*2020Distributionally}, which currently achieves state-of-the-art debiasing performance on many benchmarks.

\begin{algorithm}[t]
\caption{Lg-Augmentation: Language-guided Diffusion-based Augmentation}
\label{algo:augmentation}
\begin{algorithmic}[1]
    \Require Subset of minority group $X_{\mathrm{minor}}$, target attribute $a_t$, pseudo-bias attribute $a_b^{\prime}$
    % $(X_{\mathrm{minor}}, Y_{\mathrm{minor}}, B_{\mathrm{minor}})$
    \Statex {\textbf{Step 1: Prompt-image similarity of original minority group}}
    \State {create a prompt $\bm{p}$ based on the class and acquired bias keyword.}
    \State {$S_{\mathrm{minor}} \gets S_{\mathrm{CLIP}}(\bm{p}, X_{\mathrm{minor}})$.}
    \Statex{\textbf{Step 2: Minority-group image generation}}
    \While{$P(a_t|a_b^{\prime}) \ne P(a_t)$}
    \State {Generate image $\bm{x}_\mathrm{gen}$ by Stable Diffusion with prompt $\bm{p}$.}
    \State {$S_{\mathrm{gen}} \gets S_{\mathrm{CLIP}}(\bm{w}, \{\bm{x}_\mathrm{gen}\})$.}
    \If{$S_{\mathrm{gen}} > S_{\mathrm{minor}}$}
    \State {Add $\bm{x}_\mathrm{gen}$ to the training data.} 
    \State {Update $a_t$ and $a_b^{\prime}$.} 
    \EndIf
    \EndWhile
    \Statex{\textbf{Step 3: Train unbiased classifier}}
    \State {Train classifier on minority-augmented dataset via ERM.}
    % \newline \Return final model $f_\theta$ from Step 3.
\end{algorithmic}
\end{algorithm}

\textbf{Lg-Augmentation: Language-guided Diffusion-based Augmentation.}
This debiasing method aims to mitigate dataset bias itself by increasing data for minority groups, and is based on data augmentation by Stable Diffusion~\citep{rombach2022highresolution}.
From the linguistic nature of biases extracted by our framework, we can create prompts to describe minority groups using the corresponding target classes and bias keywords.
Our language-guided diffusion-based augmentation method (Lg-Augmentation) is as follows: (\cf~Algorithm~\ref{algo:augmentation}). 

Step 1: Prompt-image similarity of original minority group.
We first construct a prompt $\bm{p}\in\mathbb{R}^d$ based on the target class and extracted bias keywords corresponding to minority group, such as ``a photo of a \{Class\} in \{Attribute\}''. 
We then compute CLIP text-image similarity defined in Eq. (\ref{eq:score_similarity}) between the prompt and subset of minority group $X_{\mathrm{minor}}\subset\mathbb{R}^{H\times W\times 3}$. 
The computed similarity score $S_{\mathrm{minor}}\in\mathbb{R}$ is used as a threshold in Step 2.

Step 2: Minority-group image generation.
We generate images $\bm{x}_\mathrm{gen}\in\mathbb{R}^{H\times W\times 3}$ of minority group using the constructed prompt $\bm{p}$ by Stable Diffusion~\citep{rombach2022highresolution}. We then compute CLIP text-image similarity 
% (equation (~\ref{eq:score_similarity})) 
between the prompt and generated image.
If the computed similarity score $S_{\mathrm{gen}}\in\mathbb{R}$ is greater than the threshold $S_{\mathrm{minor}}$, the generated image is considered following the prompt and added to the training data.
Building upon the insight from~\citep{qraitem2023bias}, this generation process is iteratively performed until ensuring that $P(a_t|a_b^{\prime}) = P(a_t)$, meaning the uniform bias distribution across all classes.
Note that we assume a scenario where the bias attribute $a_b$ is not available. 
Instead, we use a pseudo-bias attribute $a_b^{\prime}$ obtained through pseudo-annotation with Lg-DRO.

Step 3: Train unbiased classifier.
We finally train an unbiased classifier on the minority-augmented dataset.
In the training on this debiased dataset, we employ empirical risk minimization (ERM) which minimizes the average loss over the training data, rather than DRO which is only effective in a biased dataset.
% とりあえずERM数式だけ
% \begin{equation}
%     \mathcal{L}_{\mathrm{avg}}(f_{\theta}) \coloneqq \mathbb{E}_{(x,y,a) \sim P}[\ell(f_{\theta}(x), y)].
%     \label{eq:erm} 
% \end{equation}

\textbf{Remark:} 
One might notice that existing methods have already proven that generated images can be beneficial enough to be added in training data. 
Particularly, ~\citep{dunlap2023diversify} has demonstrated that image generation by leveraging captions is able to diversify the training data, achieving high generalization performance in various tasks. 
However, note that in our framework, bias keywords are preliminarily identified by our language-guided bias detection.
Augmenting minority groups using captions might inadvertently introduce unexpected biases since captions contain various attributes beyond just the target class and bias attribute in question. 
Therefore, we opt for a straightforward prompt such as ``a photo of a \{Class\} in \{Attribute\}'' to avoid the emergence of unintended spurious correlations.

\begin{table}[t]
  \caption{\textbf{Main results on CMNIST, Waterbirds and CelebA dataset.} 
  \textbf{1st} / \underline{2nd} accuracies are \textbf{bolded} / \underline{underlined} for each dataset in the same settings.
  Vanilla refers to the baseline model without any debiasing procedures. Prior indicates prior knowledge. UA and BC are the abbreviations for unbiased accuracy (\%) and bias-conflict (\%). We find that our methods outperform existing methods and are comparable even to Group-DRO with prior knowledge.}
  % Note that Group-DRO is the only method that prior knowledge and labels of bias attribute are available.}
  \label{table:accuracy}
  \centering
  {\tabcolsep = 3.0mm
  \begin{tabular}{lcccccc}
    \toprule
    & 
    &\multicolumn{1}{c}{\textbf{CMNIST}} & \multicolumn{2}{c}{\textbf{Waterbirds}} & \multicolumn{2}{c}{\textbf{CelebA}} \\ \cmidrule(l){3-3} \cmidrule(l){4-5} \cmidrule(l){6-7} 
    \textbf{Method} & \textbf{Prior}  & \textbf{UA} & \textbf{UA} & \textbf{BC} & \textbf{UA} & \textbf{BC} \\
    \midrule
    Vanilla & \ding{55} & 10.27 & 69.83 & 34.31 & 70.25 & 53.52 \\
    LfF~\citep{nam2020learning} & \ding{55} & 85.39 & 91.20 & 80.81 & 84.24 & 81.24 \\
    JTT~\citep{liu2021just} & \ding{55} & 89.74 & 90.63 & 85.99 & 88.10 & 80.76 \\
    BPA~\citep{Seo_2022_CVPR} & \ding{55} & 85.26 & 87.05 & 71.39 & 90.18 & 82.54 \\
    CNC~\citep{zhang2022correctncontrast} & \ding{55} & 88.85 & 90.95 & 88.50 & 89.90 & \underline{88.80} \\
    GALS~\citep{9880269} & \ding{55} & 33.62 & 89.15 & 75.39 & 83.09 & 75.48 \\
    B2T-DRO~\citep{kim2024discovering} & \ding{55} & 94.44 & 91.42 & \underline{88.94} & 90.60 & 87.24 \\
    \rowcolor[gray]{0.90}
    Lg-DRO (Ours) & \ding{55} & \underline{95.37} & \bf{92.89} & \bf{89.58} & \underline{91.50} & \bf{88.89} \\
    \rowcolor[gray]{0.90}
    Lg-Augmentation (Ours) & \ding{55} & \bf{96.83} & \underline{92.28} & 87.34 & \bf{92.76} & 86.80 \\
    \midrule
    Group-DRO~\citep{Sagawa*2020Distributionally} & \ding{51} & {\color{gray} 95.81} & {\color{gray}92.51} & {\color{gray} 88.65} & {\color{gray} 92.86} & {\color{gray} 91.67} \\
    % EnD~\citep{Tartaglione_2021} & \ding{51} & 81.46 &&& 91.21 & 87.45 \\
    \bottomrule
  \end{tabular}
  }
\end{table}

\section{Experiments}
\label{sec:experiment}

\subsection{Experimental detail}
\textbf{Dataset.}
We use three datasets following existing works~\citep{Seo_2022_CVPR, nam2020learning, zhang2022correctncontrast, liu2021just}. (i) \textbf{CMNIST}~\citep{arjovsky2020invariant}:
We use CMNIST (Colored-MNIST) dataset, a extension of MNIST~\citep{mnist} with color attributes, where different bias colors are assigned to each class.
% , as shown in Figure~\ref{fig:bias_cmnist}. 
We set the proportion of images with assigned bias colors to 95\% for each class and trained a 10-class classification model for digits 0 through 9, namely $(a_t, a_b) = (\mathrm{Digit}, \mathrm{Color})$.
Note that in CMNIST, the color is applied to the digits themselves rather than the background, making it impossible to separate the class and the bias attribute within the images.
(ii) \textbf{Waterbirds}~\citep{Sagawa*2020Distributionally}:
Waterbirds dataset is created by synthesizing images of waterbirds and landbirds with water and land backgrounds. 
Given the significant correlation between classes (waterbird / landbird) and backgrounds (water / land), images belonging to atypical groups (\eg, landbird with water background) are markedly scarce. In this experiment, we train a model to classify waterbirds versus landbirds; \ie, $(a_t, a_b) = (\mathrm{Species}, \mathrm{Background})$.
(iii) \textbf{CelebA}~\citep{liu2015faceattributes}:
CelebA is a facial image dataset annotated with attributes such as gender, hair color, and expressions. Due to the strong correlation between hair color and gender, the number of blonde-male images is extremely low. We train a model to classify hair color (blonde versus non-blonde) in this section, namely $(a_t, a_b) = (\mathrm{Haircolor}, \mathrm{Gender})$.
More detailed information about the imbalance in the datasets is in Appendix~\ref{appendix:detaset_details}.

\textbf{Model.} We train ImageNet-pretrained ResNet-50~\citep{7780459} for all debiasing methods following existing work~\citep{liu2021just, zhang2022correctncontrast, 9880269, kim2024discovering}. 
We employ BLIP~\citep{li2022blip} as the VLM to generate captions for each training image. 
For language-guided bias detection, we extract five keywords with the top five highest class-specificity scores for CelebA and Waterbirds datasets, and one keyword with the highest class-specificity score for CMNIST.
These keywords are then used for pseudo-bias annotation in Lg-DRO.

\textbf{Comparision methods.}  
We compare our proposed debiasing methods with recent state-of-the-art methods ~\citep{nam2020learning, liu2021just, Seo_2022_CVPR, zhang2022correctncontrast, 9880269, kim2024discovering} on the unknown bias scenario. 
Additionally, we evaluate our methods against Group-DRO~\citep{Sagawa*2020Distributionally} that utilizes annotations of bias attributes for each dataset.
Specifically, B2T-DRO~\citep{kim2024discovering} applies the pseudo-labels of bias keywords extracted by B2T to Group-DRO.
Comparing B2T-DRO with our Lg-DRO allows us to validate the effectiveness of our keyword extraction leveraging GPT-4.
Moreover, the effectiveness of our pseudo-bias annotation is also validated by comparing Lg-DRO using pseudo-labels and Group-DRO using ground truth labels.

\textbf{Evaluation metrics.}
The evaluation metrics for the proposed method include Unbiased Accuracy (UA)~\citep{bahng2020learning, nam2020learning} and Bias-Conflict (BC)~\citep{NEURIPS2021_de8aa43e}, which are commonly used in the community ~\citep{Tartaglione_2021, Seo_2022_CVPR, 9880269, qraitem2023bias}. 
UA represents the average group accuracy, with groups divided by target classes and bias attributes, while BC represents the worst-group accuracy within all groups.
The details on the experimental settings such as hyperparameters are provided in the Appendix~\ref{appendix:settings}.

% \begin{table}[t]
%   \caption{Extracted bias keyword in CMNIST dataset.}
%   \label{table:extracted_attributes_cmnist}
%   \centering
%   % \begin{tabular}{@{}lllllllllll@{}}
%   \begin{tabular}{@{}ccccccccccc@{}}
%     \toprule
%     Class & 0 & 1 & 2 & 3 & 4 & 5 & 6 & 7 & 8 & 9 \\
%     \midrule
%     Bias Keyword & red & green & yellow & blue & orange & purple & blue & pink & yellow & red \\
%     \bottomrule
%   \end{tabular}
% \end{table}

\subsection{Main results}
\label{sec:4.2}
Experimental results of bias mitigation accuracy are presented in Table~\ref{table:accuracy}. 
The highest accuracy is indicated in bold, and the second highest accuracy is underlined in the table. 
\textit{Vanilla} represents the baseline model alone without applying any bias mitigation methods. 
Our findings are as follows:
% The results show that our debiasing methods generally outperform all existing methods that do not use prior knowledge for all datasets. 
% Especially noteworthy is the fact that our method on the CMNIST dataset notably exhibits superior performance to GALS, which also utilizes VLMs.
% This discrepancy stems from the fact that CMNIST assigns a visually indivisible bias attribute, color, to the target class digits. 
% This result demonstrates that our method effectively addresses the limitations of GALS.
% Moreover, our methods demonstrate competitiveness even when compared to Group-DRO with prior knowledge.
% Specifically, Lg-DRO surpasses Group-DRO on Waterbirds dataset in terms of both unbiased accuracy and bias-conflict.
% See Appendix~\ref{appendix:analysis} for further analysis on the experimental results.

\textbf{Our debiasing methods outperform all existing methods without prior knowledge for all datasets.}
Both Lg-DRO and Lg-Augmentation outperform all approaches without bias information in terms of unbiased accuracy, in particular, Lg-DRO surpasses all methods in both unbiased accuracy and bias-conflict.
Especially noteworthy is the fact that our method on the CMNIST dataset notably exhibits superior performance to GALS which also utilizes VLMs.
This discrepancy stems from the fact that CMNIST assigns a visually indivisible bias attribute, color, to the target class digits, demonstrating that our method effectively addresses the limitations of GALS.
Furthermore, the consistent superiority of Lg-DRO over B2T-DRO suggests that our language-guided bias detection leveraging GPT-4 can achieve more effective bias extraction.
The specific results of bias extraction are presented in Section~\ref{sec:analysis}.

\textbf{Our methods are comparable even to Group-DRO with prior knowledge.}
Both Lg-DRO and Lg-Augmentation achieve competitive unbiased accuracy across all datasets compared to Group-DRO.
Notably, Lg-DRO achieves superior performance compared to Group-DRO on Waterbirds dataset, excelling in both unbiased accuracy and bias-conflict, while Lg-Augmentation also outperforms on CMNIST dataset.

\begin{table}[t]
  \caption{\textbf{Results with ResNet-18 and ViT-small} on CelebA dataset. Our methods are comparable to Group-DRO with other backbones.
  % \textbf{1st} / \underline{2nd} accuracies are \textbf{bolded} / \underline{underlined} for each dataset in the same settings.
  % Prior indicate prior knowledge. UA and BC are the abbreviations for unbiased accuracy (\%) and bias-conflict (\%). 
  % We find that our methods outperform existing methods and are comparable even to Group-DRO with prior knowledge.}
  }
  \label{table:accuracy_additional}
  \centering
  {\tabcolsep = 3.0mm
  \begin{tabular}{lccccc}
    \toprule
    & 
    &\multicolumn{2}{c}{\textbf{ResNet-18}} &\multicolumn{2}{c}{\textbf{ViT-small}} \\ \cmidrule(l){3-4} \cmidrule(l){5-6}
    \textbf{Method} & \textbf{Prior}  &\textbf{UA} & \textbf{BC} &\textbf{UA} & \textbf{BC} \\
    \midrule
    \rowcolor[gray]{0.90}
    Lg-DRO (Ours) & \ding{55} & 90.95 & 85.56 & 91.68 & 87.78 \\
    \rowcolor[gray]{0.90}
    Lg-Augmentation (Ours) & \ding{55} & 91.43 & 83.33 & 92.01 & 86.11 \\
    \midrule
    Group-DRO~\citep{Sagawa*2020Distributionally} & \ding{51} & {\color{gray}91.83} & {\color{gray}87.78} & {\color{gray}92.21} & {\color{gray}88.89}\\
    \bottomrule
  \end{tabular}
  }
\end{table}

\subsection{Additional experiment}
We conduct additional experiments with ResNet-18~\citep{7780459} and ViT-small~\citep{touvron2021training} on CelebA dataset. 
The experimental results in Table~\ref{table:accuracy_additional} demonstrate that our methods achieve performance competitive with Group-DRO, particularly in terms of unbiased accuracy, even when applied to other models.

\begin{table}[t]
  \caption{\textbf{Comparison of extracted bias keywords} in Waterbirds and CelebA datasets. {\color{red}{Red}}
  keywords indicate inappropriate keywords as biases, such as those irrelevant or common across the classes.}
  \label{table:extracted_attributes_waterbird_celeba}
  \centering
  % \begin{tabular}{@{}cccc@{}}
  \begin{tabular}{@{}ccc@{}}
    \toprule
    \textbf{Class} & \textbf{Ours} & \textbf{B2T}~\citep{kim2024discovering} \\
    \midrule
    Waterbird & beach, lake, water, seagull, pond & beach, seagull, water, {\color{red}{tail}}, {\color{red}{dog}}\\
    \midrule
    Landbird & forest, woods, rainforest, tree branch, tree & forest, garden, woods, trees, tree\\ 
    \midrule
    Blonde & 
    \begin{tabular}{l}
    woman with blonde hair, blonde hair,\\ actress, model, woman with long hair
    \end{tabular} & 
    \begin{tabular}{l}
    actress, hair color, model,\\ long hair, {\color{red}{premiere of romantic}}
    \end{tabular} \\
    \midrule
    Non-blonde & 
      \begin{tabular}{l}
      man, man wearing sunglasses,\\ young man, black hair, actor
      \end{tabular} & 
      \begin{tabular}{l}
      man, player, comedy, artist, {\color{red}{person}}
      \end{tabular} \\
    \bottomrule
  \end{tabular}
\end{table}

\begin{table}[t]
  \caption{\textbf{Ablation on our keyword validation} (w/o Eq. (\ref{eq:score_specificity})) in Waterbirds and CelebA datasets. Our detection framework without validation process extracts invalid keywords (colored {\color{red}{red}}) as biases such as words common across the classes, suggesting the effectiveness of our keyword validation.}
  \label{table:extracted_attributes_ablation}
  \centering
  % \begin{tabular}{@{}cccc@{}}
  \begin{tabular}{@{}cc@{}}
    \toprule
     \textbf{Class} & \textbf{Extracted Bias Keywords} (w/o Eq. (\ref{eq:score_specificity})) \\
    \midrule
      Waterbird & {\color{red}{biological species}}, {\color{red}{bird}}, beach\\
      Landbird & {\color{red}{biological species}}, {\color{red}{bird}}, tree\\ 
    \midrule
      Blonde & actor, {\color{red}{person}}, {\color{red}{hair}}\\
      Non-blonde & actor, {\color{red}{person}}, {\color{red}{premiere}}\\
    \bottomrule
  \end{tabular}
\end{table}

\section{Framework analysis}
\label{sec:analysis}
\subsection{Extraction of bias keywords}
We conduct experiments to determine whether our bias detection framework can appropriately extract potential biases present in the dataset. 
Additionally, we compare bias keywords extracted by our language-guided bias detection leveraging GPT-4 and those extracted by B2T~\citep{kim2024discovering} on Waterbirds and CelebA datasets.
Furthermore, we confirm the effectiveness of our validation based on the class-specificity scores (Eq. (\ref{eq:score_specificity})) by ablation studies performing the same experiments without the validation process. 
% For the encoders used to verify bias attributes, we utilized the image and text encoders from CLIP~\citep{radford2021learning}.

\begin{figure}[t]
  \centering
  \includegraphics[keepaspectratio, scale=0.4]{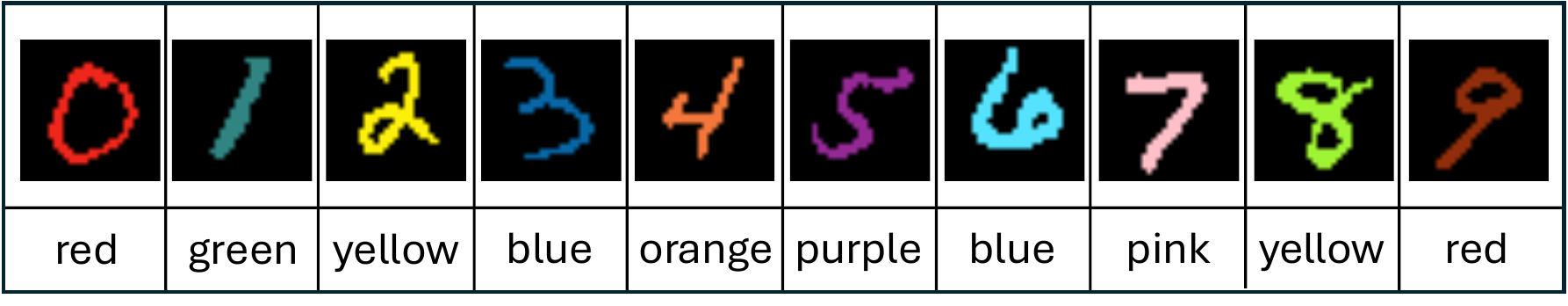}
  \caption{samples with bias-aligned colors and extracted bias keywords in CMNIST dataset.}
  \label{fig:bias_cmnist}
\end{figure}

The results of the bias detection and comparison with B2T are shown in Figure~\ref{fig:bias_cmnist} and Table~\ref{table:extracted_attributes_waterbird_celeba}. 
In CMNIST dataset, the extracted biases for each class are generally aligned with the bias colors. 
Furthermore, in Waterbirds dataset, background information is extracted as bias keywords, whereas in CelebA dataset, gender-related words are identified as biases, both consistent with biases problematic in each dataset. 
Remarkably, the keywords extracted by B2T include common words across the classes (\eg ``person'' in CelebA dataset) and irrelevant words (\eg ``dog'' in Waterbirds dataset) as biases, suggesting that our detection framework achieves more precise bias identification.
When validation based on class-specificity scores is omitted (shown in Table~\ref{table:extracted_attributes_ablation}), attributes common to both classes (\eg ``bird'' in Waterbirds dataset and ``person'' in CelebA dataset) are erroneously identified as bias attributes. 
This indicates that validation based on class-specificity scores is effective for accurate performance of our bias detection framework.

\subsection{Visualization of generated images}
\begin{figure*}[t]
\centering
    \includegraphics[width=0.85\linewidth]{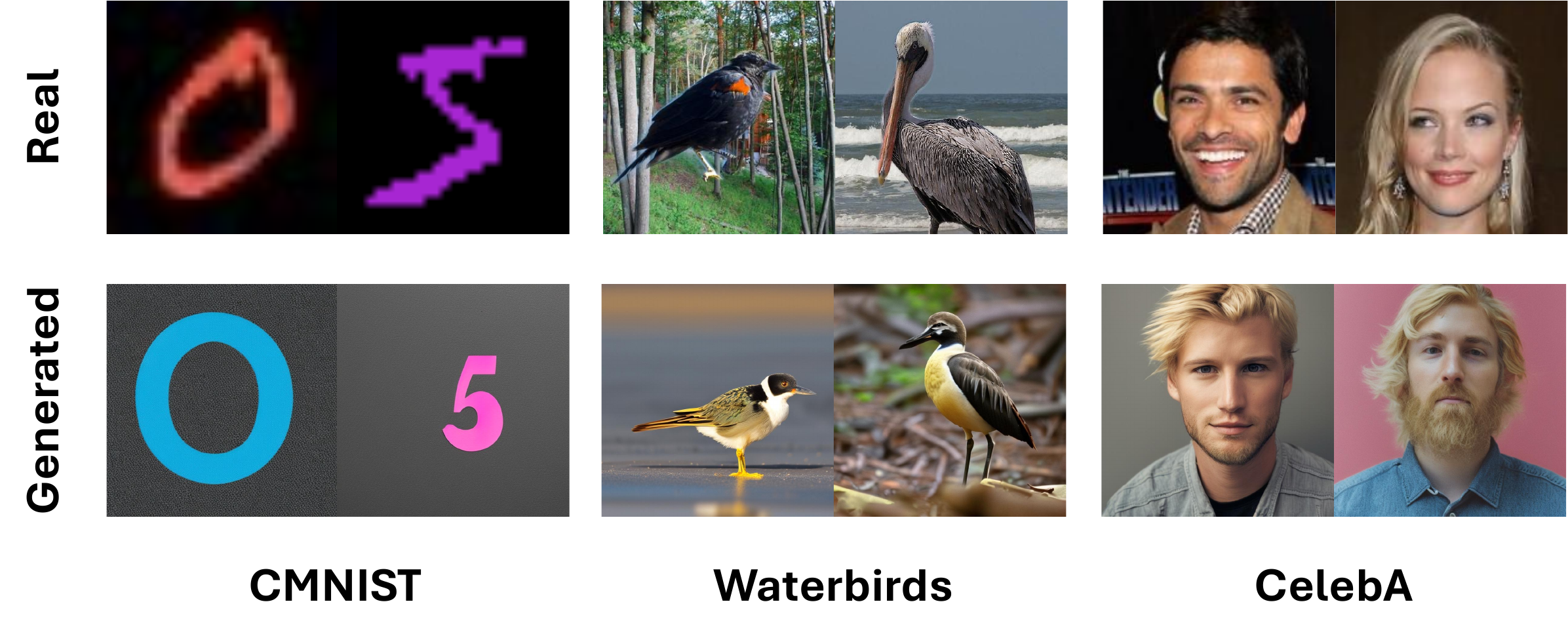}\\
    \vspace{-5pt}
    \caption{\textbf{Visualization of generated images by our Language-guided Diffusion-based Augmentation.} The generated images are consistent with minority groups (\eg, landbird on water, blonde man), and effective enough for resolving data imbalance.}
    \label{fig:samples_generated}
\end{figure*}

Visualization of generated images and real images from each dataset are presented in Figure~\ref{fig:samples_generated}. 
We use prompts describing detected minority groups as input for image generation. 
Namely, images of digit with bias-conflict colors for CMNIST, bird in atypical background for Waterbirds and blonde man for CelebA are generated by Stable Diffusion.
The images illustrated in Figure~\ref{fig:samples_generated} are indeed consistent with the input prompts, sufficiently augmenting the data for minority groups.
Additional visualization of generated images can be found in Appendix~\ref{appendix:visualization}.

\section{Conclusion}
\label{sec:conclusion}
In this study, we presented a language-guided framework for unknown bias detection that extracts potential dataset biases as keywords, making these biases more interpretable.
We further proposed two debiasing methods: (a) leveraging Group-DRO through pseudo-annotation of detected bias, and (b) data augmentation via Stable Diffusion, leveraging acquired bias keywords as prompts. 
We conducted experiments on CMNIST, Waterbirds, and CelebA datasets.
The experimental results show that our methods outperform existing methods without prior knowledge and are even competitive with Group-DRO that requires prior knowledge, despite being designed for scenarios where biases are unknown and bias labels are unavailable.
Given that our detection framework presents biases as keywords, it is particularly suitable for real-world applications, offering high interpretability and effective bias mitigation performance.

\textbf{Limitations.}
While our framework identifies biases from captions generated by VLM, it cannot be applied to attributes that VLMs cannot represent. 
For instance, current VLMs seldom express hair length in captions, making it challenging to detect biases based on gender and hair length (\eg, short-haired women). 
Although it is possible to prompt the model to describe hair length through captions, this process requires some prior knowledge. 
Despite our experiments have demonstrated that our framework can detect and mitigate novel biases, it may not be capable of identifying all biases with the same effectiveness.

\textbf{Broader impact.}
Dataset bias is a well-documented concern in society, and utilizing models trained on biased datasets without acknowledging the presence of bias could yield controversial results in real-world applications, especially when the bias is related to sensitive attributes such as race or gender.
% For example, it has been demonstrated that commercial face recognition models exhibit significantly higher misclassification rates for black women compared to white men~\cite{birhane2021multimodal}.
% Moreover, classification models have been observed to amplify existing societal biases~\cite{wang2019balanced}, making bias an undeniable challenge for the wider societal acceptance of classification models.
In this paper, we proposed a framework to identify and mitigate potential dataset biases to reduce such risks, aiming to achieve a positive societal impact.
% Additionally, 
% while this paper focuses on training unbiased classifiers for evaluations of our framework, 
% our framework targets the dataset itself rather than the model, suggesting its broader applicability across various tasks.
% For instance, while generative models have attracted significant attention in recent years, concerns persist regarding biases in the generated images. 
% Our framework offers the potential to generate more balanced data as shown in Figure~\ref{fig:diffusion_application} by incorporating detailed prompts, including bias attributes, through the extraction of unknown biases as keywords.
It is conceivable that as the performance of VLMs improves, our framework will be able to mitigate an increasing range of biases. 
It is equally important, however, to recognize the ongoing debate regarding the potential risks of VLMs to amplify biases, as the issues and mitigation strategies addressed in~\citep{10.1007/978-3-030-01219-9_47, hirota2023modelagnostic}.

\newpage
\bibliographystyle{unsrt}
\bibliography{cite}

\begin{thebibliography}{10}

\bibitem{Lapuschkin_2019}
Sebastian Lapuschkin, Stephan Wäldchen, Alexander Binder, Grégoire Montavon, Wojciech Samek, and Klaus-Robert Müller.
\newblock Unmasking clever hans predictors and assessing what machines really learn.
\newblock {\em Nature Communications}, 10(1), 2019.

\bibitem{wang2019balanced}
Tianlu Wang, Jieyu Zhao, Mark Yatskar, Kai-Wei Chang, and Vicente Ordonez.
\newblock Balanced datasets are not enough: Estimating and mitigating gender bias in deep image representations.
\newblock In {\em International Conference on Computer Vision (ICCV)}, 2019.

\bibitem{ramaswamy2021fair}
Vikram~V. Ramaswamy, Sunnie S.~Y. Kim, and Olga Russakovsky.
\newblock Fair attribute classification through latent space de-biasing.
\newblock In {\em Conference on Computer Vision and Pattern Recognition (CVPR)}, 2021.

\bibitem{li2019repair}
Yi~Li and Nuno Vasconcelos.
\newblock Repair: Removing representation bias by dataset resampling.
\newblock In {\em Conference on Computer Vision and Pattern Recognition (CVPR)}, 2019.

\bibitem{byrd2019effect}
Jonathon Byrd and Zachary~C. Lipton.
\newblock What is the effect of importance weighting in deep learning?
\newblock In {\em International Conference on Machine Learning (ICML)}, 2019.

\bibitem{qraitem2023bias}
Maan Qraitem, Kate Saenko, and Bryan~A. Plummer.
\newblock Bias mimicking: A simple sampling approach for bias mitigation.
\newblock In {\em Conference on Computer Vision and Pattern Recognition (CVPR)}, 2023.

\bibitem{wang2020fairness}
Zeyu Wang, Klint Qinami, Ioannis~Christos Karakozis, Kyle Genova, Prem Nair, Kenji Hata, and Olga Russakovsky.
\newblock Towards fairness in visual recognition: Effective strategies for bias mitigation.
\newblock In {\em Conference on Computer Vision and Pattern Recognition (CVPR)}, 2020.

\bibitem{10.5555/1293951.1293954}
Nathalie Japkowicz and Shaju Stephen.
\newblock The class imbalance problem: A systematic study.
\newblock {\em Intell. Data Anal.}, 6(5):429–449, 2002.

\bibitem{Sagawa*2020Distributionally}
Shiori Sagawa, Pang~Wei Koh, Tatsunori~B. Hashimoto, and Percy Liang.
\newblock Distributionally robust neural networks for group shifts: On the importance of regularization for worst-case generalization.
\newblock In {\em International Conference on Learning Representations (ICLR)}, 2020.

\bibitem{ross2017right}
Andrew~Slavin Ross, Michael~C. Hughes, and Finale Doshi-Velez.
\newblock Right for the right reasons: Training differentiable models by constraining their explanations.
\newblock In {\em International Joint Conferences on Artificial Intelligence (IJCAI)}, 2017.

\bibitem{alvi2018turning}
Mohsan Alvi, Andrew Zisserman, and Christoffer Nellaker.
\newblock Turning a blind eye: Explicit removal of biases and variation from deep neural network embeddings.
\newblock In {\em European Conference on Computer Vision (ECCV), Workshop on Bias Estimation in Face Analytics}, 2018.

\bibitem{Tartaglione_2021}
Enzo Tartaglione, Carlo~Alberto Barbano, and Marco Grangetto.
\newblock End: Entangling and disentangling deep representations for bias correction.
\newblock In {\em Conference on Computer Vision and Pattern Recognition (CVPR)}, 2021.

\bibitem{kim2019learning}
Byungju Kim, Hyunwoo Kim, Kyungsu Kim, Sungjin Kim, and Junmo Kim.
\newblock Learning not to learn: Training deep neural networks with biased data.
\newblock In {\em Conference on Computer Vision and Pattern Recognition (CVPR)}, 2019.

\bibitem{NEURIPS2021_de8aa43e}
Youngkyu Hong and Eunho Yang.
\newblock Unbiased classification through bias-contrastive and bias-balanced learning.
\newblock In {\em Conference on Neural Information Processing Systems (NeurIPS)}, 2021.

\bibitem{nam2020learning}
Junhyun Nam, Hyuntak Cha, Sungsoo Ahn, Jaeho Lee, and Jinwoo Shin.
\newblock Learning from failure: Training debiased classifier from biased classifier.
\newblock In {\em Conference on Neural Information Processing Systems (NeurIPS)}, 2020.

\bibitem{liu2021just}
Evan~Zheran Liu, Behzad Haghgoo, Annie~S. Chen, Aditi Raghunathan, Pang~Wei Koh, Shiori Sagawa, Percy Liang, and Chelsea Finn.
\newblock Just train twice: Improving group robustness without training group information.
\newblock In {\em International Conference on Machine Learning (ICML)}, 2021.

\bibitem{Seo_2022_CVPR}
Seonguk Seo, Joon-Young Lee, and Bohyung Han.
\newblock Unsupervised learning of debiased representations with pseudo-attributes.
\newblock In {\em Conference on Computer Vision and Pattern Recognition (CVPR)}, 2022.

\bibitem{zhang2022correctncontrast}
Michael Zhang, Nimit~S. Sohoni, Hongyang~R. Zhang, Chelsea Finn, and Christopher Ré.
\newblock Correct-n-contrast: A contrastive approach for improving robustness to spurious correlations.
\newblock In {\em International Conference on Machine Learning (ICML)}, 2022.

\bibitem{9880269}
Suzanne Petryk, Lisa Dunlap, Keyan Nasseri, Joseph Gonzalez, Trevor Darrell, and Anna Rohrbach.
\newblock On guiding visual attention with language specification.
\newblock In {\em Conference on Computer Vision and Pattern Recognition (CVPR)}, 2022.

\bibitem{kim2024discovering}
Younghyun Kim, Sangwoo Mo, Minkyu Kim, Kyungmin Lee, Jaeho Lee, and Jinwoo Shin.
\newblock Discovering and mitigating visual biases through keyword explanation.
\newblock In {\em Conference on Computer Vision and Pattern Recognition (CVPR)}, 2024.

\bibitem{dosovitskiy2021image}
Alexey Dosovitskiy, Lucas Beyer, Alexander Kolesnikov, Dirk Weissenborn, Xiaohua Zhai, Thomas Unterthiner, Mostafa Dehghani, Matthias Minderer, Georg Heigold, Sylvain Gelly, Jakob Uszkoreit, and Neil Houlsby.
\newblock An image is worth 16x16 words: Transformers for image recognition at scale.
\newblock In {\em International Conference on Learning Representations (ICLR)}, 2021.

\bibitem{CAMPOS2020257}
Ricardo Campos, Vítor Mangaravite, Arian Pasquali, Alípio Jorge, Célia Nunes, and Adam Jatowt.
\newblock Yake! keyword extraction from single documents using multiple local features.
\newblock {\em Information Sciences}, 509:257--289, 2020.

\bibitem{10.1007/978-3-319-76941-7_63}
Ricardo Campos, V{\'i}tor Mangaravite, Arian Pasquali, Al{\'i}pio~M{\'a}rio Jorge, C{\'e}lia Nunes, and Adam Jatowt.
\newblock A text feature based automatic keyword extraction method for single documents.
\newblock In {\em European Conference on Information Retrieval (ECIR)}, 2018.

\bibitem{10.1007/978-3-319-76941-7_80}
Ricardo Campos, V{\'i}tor Mangaravite, Arian Pasquali, Al{\'i}pio~M{\'a}rio Jorge, C{\'e}lia Nunes, and Adam Jatowt.
\newblock Yake! collection-independent automatic keyword extractor.
\newblock In {\em European Conference on Information Retrieval (ECIR)}, 2018.

\bibitem{nokey}
Robin Rombach, Andreas Blattmann, Dominik Lorenz, Patrick Esser, and Bjﾃｶrn Ommer.
\newblock High-resolution image synthesis with latent diffusion models.
\newblock In {\em Conference on Computer Vision and Pattern Recognition (CVPR)}, 2022.

\bibitem{openai2024gpt4}
OpenAI.
\newblock Gpt-4 technical report.
\newblock {\em arXiv preprint arXiv:2303.08774}, 2024.

\bibitem{arjovsky2020invariant}
Martin Arjovsky, Léon Bottou, Ishaan Gulrajani, and David Lopez-Paz.
\newblock Invariant risk minimization.
\newblock {\em arXiv preprint arXiv:1907.02893}, 2020.

\bibitem{liu2015faceattributes}
Ziwei Liu, Ping Luo, Xiaogang Wang, and Xiaoou Tang.
\newblock Deep learning face attributes in the wild.
\newblock In {\em International Conference on Computer Vision (ICCV)}, 2015.

\bibitem{goodfellow2014generative}
Ian~J. Goodfellow, Jean Pouget-Abadie, Mehdi Mirza, Bing Xu, David Warde-Farley, Sherjil Ozair, Aaron Courville, and Yoshua Bengio.
\newblock Generative adversarial networks.
\newblock {\em Advances in Neural Information Processing Systems (NeurIPS)}, 27:2672--2680, 2014.

\bibitem{shafieezadehabadeh2015distributionally}
Soroosh Shafieezadeh-Abadeh, Peyman~Mohajerin Esfahani, and Daniel Kuhn.
\newblock Distributionally robust logistic regression.
\newblock In {\em Conference on Neural Information Processing Systems (NeurIPS)}, 2015.

\bibitem{hashimoto2018fairness}
Tatsunori~B. Hashimoto, Megha Srivastava, Hongseok Namkoong, and Percy Liang.
\newblock Fairness without demographics in repeated loss minimization.
\newblock In {\em International Conference on Machine Learning (ICML)}, 2018.

\bibitem{zhang2024visionlanguage}
Jingyi Zhang, Jiaxing Huang, Sheng Jin, and Shijian Lu.
\newblock Vision-language models for vision tasks: A survey.
\newblock {\em arXiv preprint arXiv:2304.00685}, 2024.

\bibitem{hu2022make}
Yaosi Hu, Chong Luo, and Zhenzhong Chen.
\newblock Make it move: Controllable image-to-video generation with text descriptions.
\newblock In {\em Conference on Computer Vision and Pattern Recognition (CVPR)}, 2022.

\bibitem{zhou2022learning}
Kaiyang Zhou, Jingkang Yang, Chen~Change Loy, and Ziwei Liu.
\newblock Learning to prompt for vision-language models.
\newblock {\em International Journal of Computer Vision (IJCV)}, 130(9):2337--2348, 2022.

\bibitem{changpinyo2022need}
Soravit Changpinyo, Doron Kukliansky, Idan Szpektor, Xi~Chen, Nan Ding, and Radu Soricut.
\newblock All you may need for vqa are image captions.
\newblock In {\em Conference of the North American Chapter of the Association for Computational Linguistics (NAACL)}, 2022.

\bibitem{miyai2024locoop}
Atsuyuki Miyai, Qing Yu, Go~Irie, and Kiyoharu Aizawa.
\newblock Locoop: Few-shot out-of-distribution detection via prompt learning.
\newblock In {\em Advances in Neural Information Processing Systems (NeurIPS)}, volume~36, 2023.

\bibitem{Avrahami_2022}
Omri Avrahami, Dani Lischinski, and Ohad Fried.
\newblock Blended diffusion for text-driven editing of natural images.
\newblock In {\em Conference on Computer Vision and Pattern Recognition (CVPR)}. IEEE, 2022.

\bibitem{NEURIPS2022_5bf2b802}
Manli Shu, Weili Nie, De-An Huang, Zhiding Yu, Tom Goldstein, Anima Anandkumar, and Chaowei Xiao.
\newblock Test-time prompt tuning for zero-shot generalization in vision-language models.
\newblock In {\em Advances in Neural Information Processing Systems (NeurIPS)}, volume~35, 2022.

\bibitem{radford2021trans}
Alec Radford, Jong~Wook Kim, Chris Hallacy, Aditya Ramesh, Gabriel Goh, Sandhini Agarwal, Girish Sastry, Amanda Askell, Pamela Mishkin, Jack Clark, et~al.
\newblock Learning transferable visual models from natural language supervision.
\newblock In {\em International Conference on Machine Learning (ICML)}, 2021.

\bibitem{rombach2022highresolution}
Robin Rombach, Andreas Blattmann, Dominik Lorenz, Patrick Esser, and Björn Ommer.
\newblock High-resolution image synthesis with latent diffusion models.
\newblock In {\em Conference on Computer Vision and Pattern Recognition (CVPR)}, 2022.

\bibitem{dunlap2023diversify}
Lisa Dunlap, Alyssa Umino, Han Zhang, Jiezhi Yang, Joseph~E. Gonzalez, and Trevor Darrell.
\newblock Diversify your vision datasets with automatic diffusion-based augmentation.
\newblock In {\em Conference on Neural Information Processing Systems (NeurIPS)}, 2023.

\bibitem{mnist}
Yann LeCun, Corinna Cortes, and Christopher J.~C. Burges.
\newblock The mnist database of handwritten digits.
\newblock http://yann.lecun.com/exdb/mnist/, 1994.

\bibitem{7780459}
Kaiming He, Xiangyu Zhang, Shaoqing Ren, and Jian Sun.
\newblock Deep residual learning for image recognition.
\newblock In {\em Conference on Computer Vision and Pattern Recognition (CVPR)}, 2016.

\bibitem{li2022blip}
Junnan Li, Dongxu Li, Caiming Xiong, and Steven Hoi.
\newblock Blip: Bootstrapping language-image pre-training for unified vision-language understanding and generation.
\newblock In {\em International Conference on Machine Learning (ICML)}, 2022.

\bibitem{bahng2020learning}
Hyojin Bahng, Sanghyuk Chun, Sangdoo Yun, Jaegul Choo, and Seong~Joon Oh.
\newblock Learning de-biased representations with biased representations.
\newblock In {\em International Conference on Machine Learning (ICML)}, 2020.

\bibitem{touvron2021training}
Hugo Touvron, Matthieu Cord, Matthijs Douze, Francisco Massa, Alexandre Sablayrolles, and Hervé Jégou.
\newblock Training data-efficient image transformers \& distillation through attention.
\newblock In {\em International Conference on Machine Learning (ICML)}, 2021.

\bibitem{10.1007/978-3-030-01219-9_47}
Lisa~Anne Hendricks, Kaylee Burns, Kate Saenko, Trevor Darrell, and Anna Rohrbach.
\newblock Women also snowboard: Overcoming bias in captioning models.
\newblock In Vittorio Ferrari, Martial Hebert, Cristian Sminchisescu, and Yair Weiss, editors, {\em European Conference on Computer Vision (ECCV)}, 2018.

\bibitem{hirota2023modelagnostic}
Yusuke Hirota, Yuta Nakashima, and Noa Garcia.
\newblock Model-agnostic gender debiased image captioning.
\newblock In {\em Conference on Computer Vision and Pattern Recognition (CVPR)}, 2023.

\end{thebibliography}

%%%%%%%%%%%%%%%%%%%%%%%%%%%%%%%%%%%%%%%%%%%%%%%%%%%%%%%%%%%%
\newpage
\appendix

\section*{Supplementary material}
This supplementary material provides dataset details (Appendix~\ref{appendix:detaset_details}), experimental details (Appendix~\ref{appendix:settings}) and additional visualization of generated images in Lg-Augmentation (Appendix~\ref{appendix:visualization})).

\section{Dataset details}
\label{appendix:detaset_details}
\begin{figure}[H]
  \centering
  \includegraphics[keepaspectratio, scale=0.35]{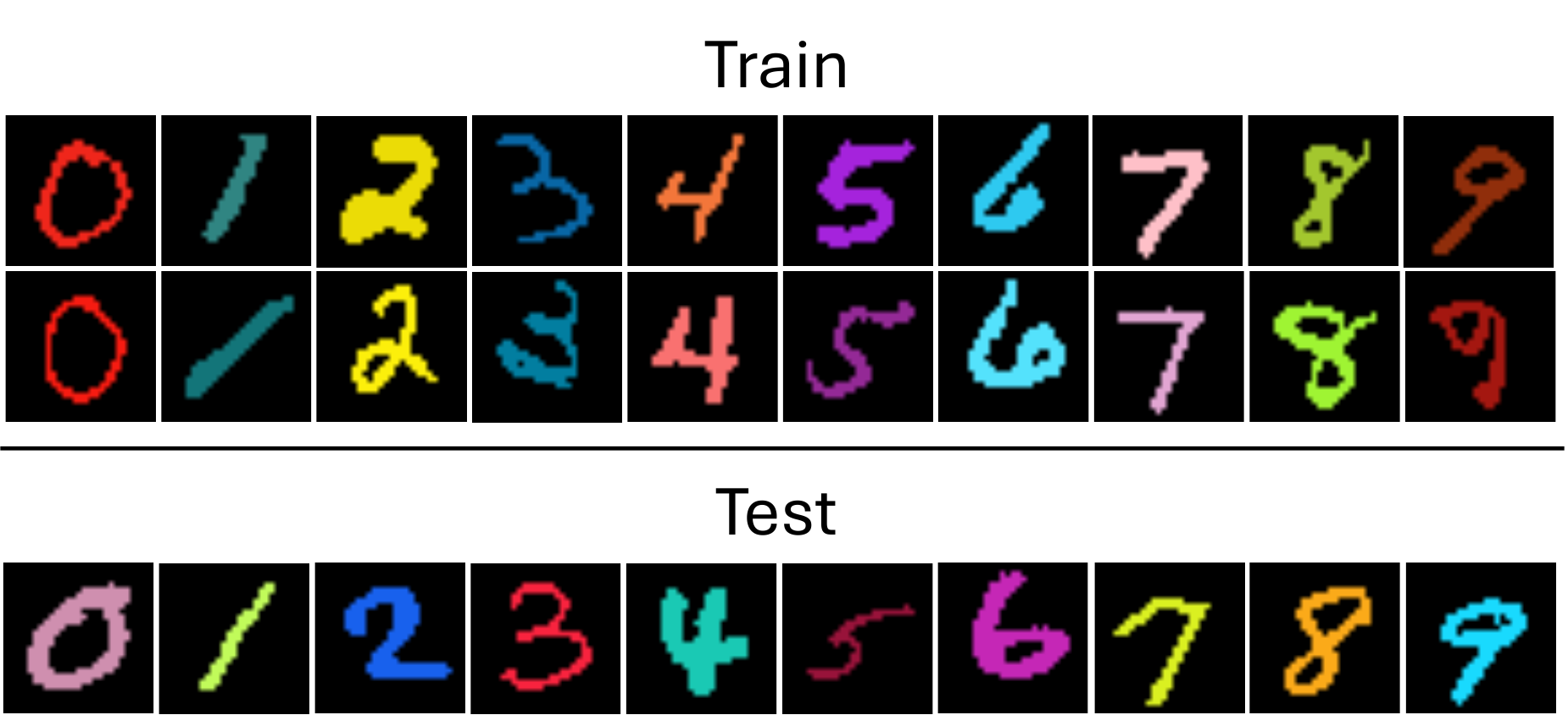}
  \caption{Samples from CMNIST training and test data.}
\label{fig:cmnist_overview}
\end{figure}

% CMNISTは特殊なのでデータセットの図作る
\textbf{CMNIST.} In CMNIST dataset, the training data has a strong color bias for each class, while the test data has a balanced color distribution as shown in Figure~\ref{fig:cmnist_overview} and Table~\ref{table:imbalance_cmnist}. We created CMNIST dataset following the code provided by the LfF repository.\footnote{\url{https://github.com/alinlab/LfF}}

\begin{table}[H]
  \caption{Imbalance of training and test data on CMNIST dataset.}
  \label{table:imbalance_cmnist}
  \centering
  \begin{tabular}{@{}ccc@{}}
    \toprule
     & Bias-aligned color & Bias-conflict color \\
    \midrule
    Train & 57,000 (95\%) & 3,000 (5\%) \\
    \midrule
    Test & 1,000 (10\%) & 9,000 (90\%) \\
    \bottomrule
  \end{tabular}
\end{table}

\textbf{Waterbirds.} The training data of Waterbirds dataset is heavily biased towards background information as shown in Table~\ref{table:imbalance_waterbirds}, and the test data exhibits the same tendency. 

\begin{table}[H]
  \caption{Imbalance of training data on Waterbirds dataset.}
  \label{table:imbalance_waterbirds}
  \centering
  \begin{tabular}{@{}ccc@{}}
    \toprule
     & Water & Land \\
    \midrule
    Waterbird & 1,057 (22\%) & 56 (1\%) \\
    \midrule
    Landbird & 184 (4\%) & 3,498 (73\%) \\
    \bottomrule
  \end{tabular}
\end{table}

\textbf{CelebA.} CelebA dataset includes annotations for 40 attributes, with a notably strong spurious correlation between hair color and gender in both training data (as presented in Table~\ref{table:imbalance_celeba}) and test data.
\begin{table}[H]
  \caption{Imbalance of training data on CelebA dataset.}
  \label{table:imbalance_celeba}
  \centering
  \begin{tabular}{@{}ccc@{}}
    \toprule
     & Male & Female \\
    \midrule
    Blonde & 1,387 (1\%) & 22,880 (14\%)\\
    \midrule
    Non-blonde & 66,874 (41\%) & 71,629 (44\%) \\
    \bottomrule
  \end{tabular}
\end{table}

\newpage

\section{Experimental settings}
\label{appendix:settings}
\subsection{Training details}
We conduct experiments using a machine with eight Tesla V100 GPUs, and each experiment utilizes four GPUs in parallel.

\textbf{CMNIST.} We train ImageNet-pretrained ResNet-50 for 50 epochs. We use batch size 512, SGD optimizer, learning rate 1e-2, momentum 0.9 and weight decay 1e-2 on both Lg-DRO and Lg-Augmentation. 

\textbf{Waterbirds.} We train ImageNet-pretrained ResNet-50 for 100 epochs. We use batch size 256, SGD optimizer, learning rate 1e-1, momentum 0.9 and weight decay 5e-3 on both Lg-DRO and Lg-Augmentation.

\textbf{CelebA.} We train ImageNet-pretrained ResNet-50, ImageNet-pretrained ResNet-18, pretrained ViT\_small\_patch16\_224 for 50 epochs. For Resnet-50 and Resnet-18, we use batch size 256, SGD optimizer, learning rate 1e-1, momentum 0.9 and weight decay 5e-3 on both Lg-DRO and Lg-Augmentation. For ViT\_small\_patch16\_224, we we use batch size 256, SGD optimizer, learning rate 1e-2, momentum 0.9 and weight decay 5e-5 for the experiments on both Lg-DRO and Lg-Augmentation.

\subsection{Language-guided bias detection}
In our keyword extraction process, we first group the generated captions by its corresponding target class. 
We then merge all the captions within each group into text and provide it to GPT-4 along with the input prompt shown in Table~\ref{table:prompt_gpt4}.

\begin{table}[H]
  \caption{Input prompt for GPT-4.}
  \label{table:prompt_gpt4}
  \centering
  \begin{tabular}{c}
    \toprule
    \rowcolor[gray]{0.90}
    You will be provided with a block of text, and your task\\
    \rowcolor[gray]{0.90}
    is to extract a list of predominant keywords from it.\\ 
    \bottomrule
  \end{tabular}
\end{table}

\subsection{Language-guided Group-DRO}
In Lg-DRO, we perform pseudo-bias annotation leveraging zero-shot classification. 
Zero-shot classification is a method based on the similarity between prompts and images.
To create prompts capable of handling a wider variety of images, we employ various prompt templates by modifying the base prompt, such as changing nouns or adding adjectives. 
Examples of the templates used in Lg-DRO are shown in Table~\ref{table:prompt_annotation}. 
The use of these prompts allows us to perform pseudo-annotation with consideration for factors such as resolution and image style.

\begin{table}[H]
  \caption{Examples of prompt templates in pseudo-bias annotation.}
  \label{table:prompt_annotation}
  \centering
  % \begin{tabular}{@{}cccc@{}}
  \begin{tabular}{@{}cc@{}}
    \toprule
     Dataset & Prompt \\
    \midrule
      CMNIST & ``a photo of a \{Attribute\} number \{Class\}'' (Base prompt)\\
      & ``a painting of a \{Attribute\} number \{Class\}''\\
      & ``a pixelated photo of a \{Attribute\} number \{Class\}''\\
      & ``a dark photo of a \{Attribute\} number \{Class\}''\\
    \midrule
      Waterbirds & ``a photo of a \{Class\} in \{Attribute\}`` (Base prompt) \\
      & ``a painting of a \{Class\} in \{Attribute\}''\\ 
      & ``a close-up photo of a \{Class\} in \{Attribute\}''\\ 
      & ``a photo of a small \{Class\} in \{Attribute\}'' \\ 
    \midrule
      CelebA & ``a photo of a \{Attribute\} with \{Class\} hair'' (Base prompt) \\
      & ``a sketch of a \{Attribute\} with \{Class\} hair'' \\
      & ``a blurry photo of a \{Attribute\} with \{Class\} hair'' \\
      & ``a black and white photo of a \{Attribute\} with \{Class\} hair'' \\ 
    \bottomrule
  \end{tabular}
\end{table}

\subsection{Language-guided Diffusion-based Augmentation}
In Lg-Augmentation, we construct prompts describing minority groups (as illustrated in Table~\ref{table:prompt_generation}) based on target class and extracted keywords.
We input these prompts into Stable Diffusion to perform data augmentation for minority groups.
The number of images generated for each dataset is shown in Table~\ref{table:augment_cmnist},~\ref{table:augment_waterbirds} and~\ref{table:augment_celeba}.

\begin{table}[H]
  \caption{Prompts used for image generation in each dataset.}
  \label{table:prompt_generation}
  \centering
  \begin{tabular}{@{}cc@{}}
    \toprule
     Dataset & Prompt \\
    \midrule
      CMNIST & colors = [``red'', ``green'', ``yellow'',  ``blue'', ``orange'',\\
      &  ``purple'', ``blue'', ``pink'', ``yellow'', ``red''] \\
      & digits = [``zero'', ``one'', ``two'', ..., ``seven'', ``eight'', ``nine''] \\
      & ``a \{colors[$i$]\} number \{digits[$j$]\} on black background'' ($0\leq i,j\leq9, i \neq j$)   \\ 
    \midrule
      Waterbirds & ``a photo of a landbird in a beach'' \\
       & ``a photo of a waterbird in a forest''\\ 
    \midrule
      CelebA & ``a photo of a man with blonde hair''\\
    \bottomrule
  \end{tabular}
\end{table}

\begin{table}[H]
  \caption{Augmented training data on CMNIST dataset. \textbf{Bolded} number indicates the number of generated images.}
  \label{table:augment_cmnist}
  \centering
  \begin{tabular}{@{}cc@{}}
    \toprule
    Bias-aligned color & Bias-conflict color \\
    \midrule
    57,000 (50\%) & 3,000 + \textbf{54,000} (50\%) \\
    \bottomrule
  \end{tabular}
\end{table}

\begin{table}[H]
  \caption{Augmented training data on Waterbirds dataset.}
  \label{table:augment_waterbirds}
  \centering
  \begin{tabular}{@{}ccc@{}}
    \toprule
     & Water & Land \\
    \midrule
    Waterbird & 1,057 (12\%) & 56 + \textbf{1,001} (12\%) \\
    \midrule
    Landbird & 184 + \textbf{3,314} (38\%) & 3,498 (38\%) \\
    \bottomrule
  \end{tabular}
\end{table}

\begin{table}[H]
  \caption{Augmented training data on CelebA dataset. Note that the number of images to generate is determined to ensure the uniform bias distribution across all classes.}
  \label{table:augment_celeba}
  \centering
  \begin{tabular}{@{}ccc@{}}
    \toprule
     & Male & Female \\
    \midrule
    Blonde & 1,387 + \textbf{19,975} (12\%) & 22,880 (13\%)\\
    \midrule
    Non-blonde & 66,874 (37\%) & 71,629 (39\%) \\
    \bottomrule
  \end{tabular}
\end{table}

\newpage

\section{Additional visualization of generated images in Lg-Augmentation}
\label{appendix:visualization}
\begin{figure}[H]
  \centering
  \includegraphics[keepaspectratio, scale=0.35]{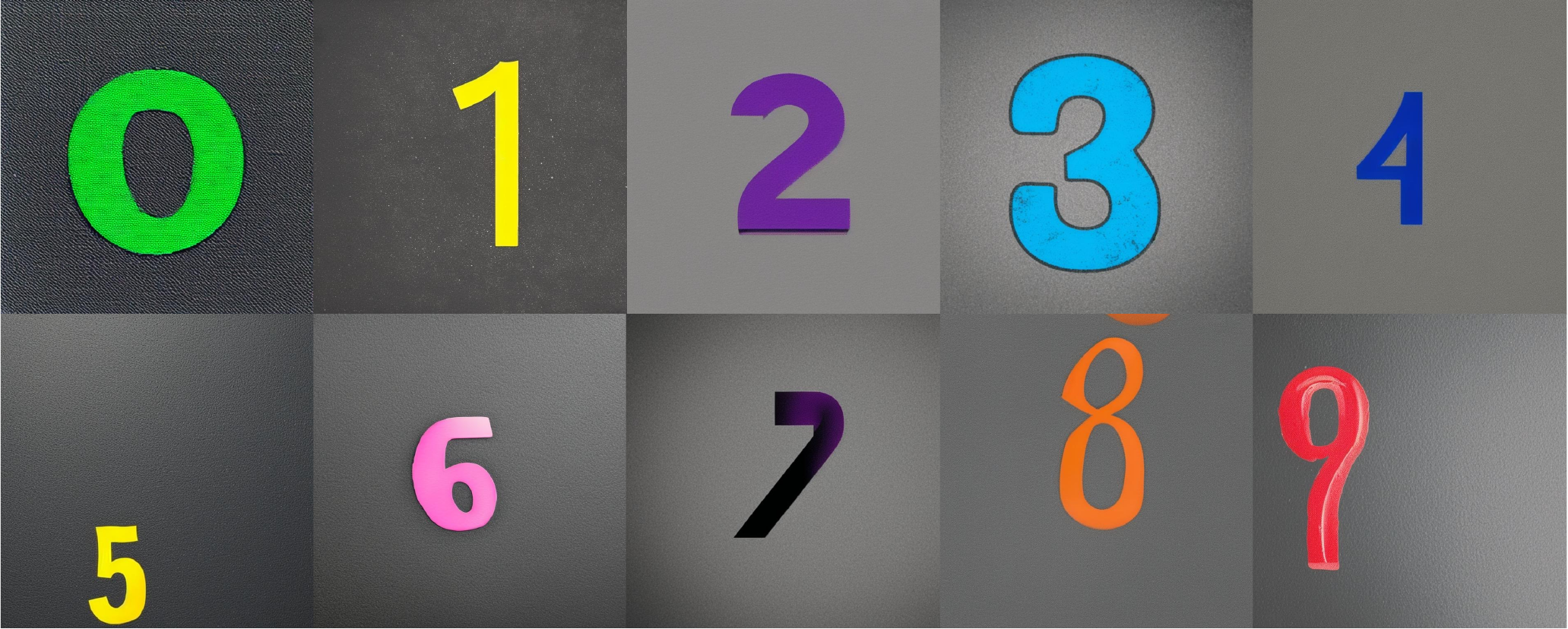}
  \caption{Additional visualization of generated images on CMNIST dataset. We generate images of digit with bias-conflict colors for each class (\eg, bias-aligned color for class zero is red, thus generating digit zero with colors other than red).}
\label{fig:cmnist_generated}
\end{figure}

\begin{figure}[H]
  \centering
  \includegraphics[keepaspectratio, scale=0.35]{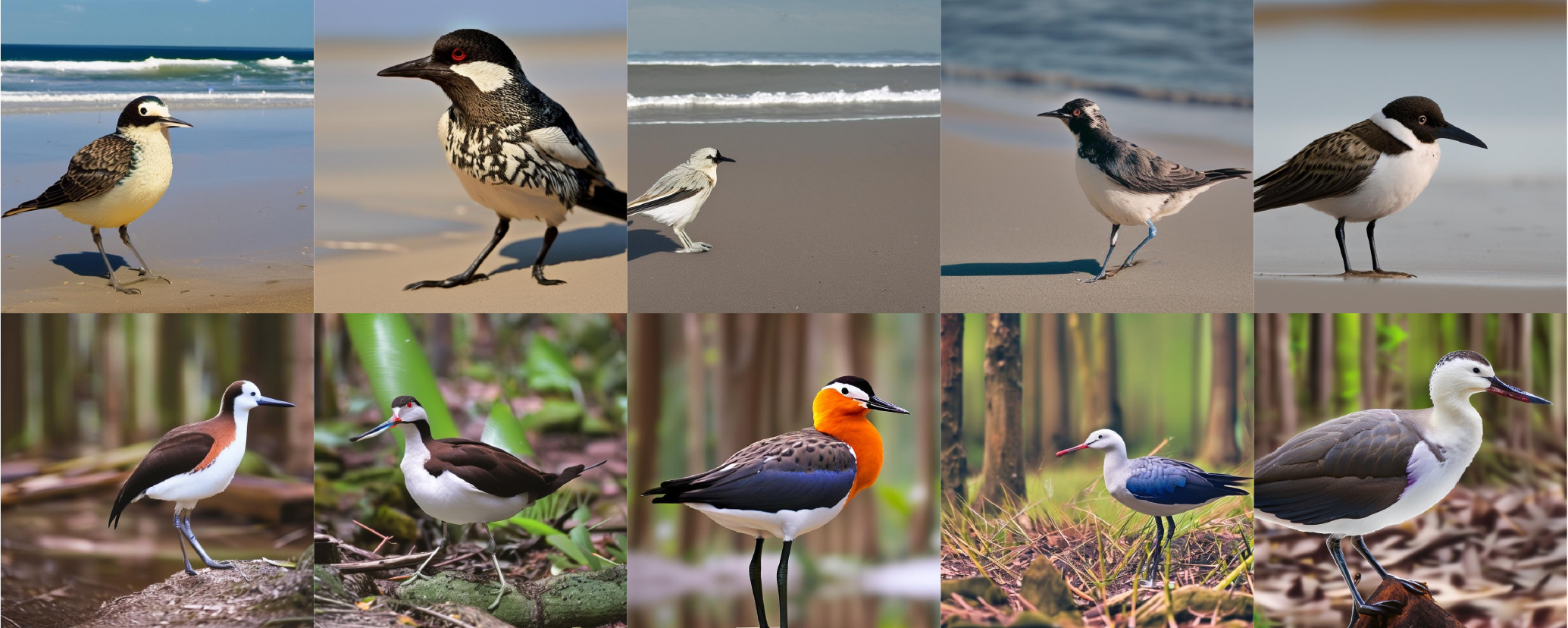}
  \caption{Additional visualization of generated images on Waterbirds dataset. We generate images of bird on atypical background for each class such as landbird with water background and waterbird with land background.}
\label{fig:waterbirds_generated}
\end{figure}

\begin{figure}[H]
  \centering
  \includegraphics[keepaspectratio, scale=0.35]{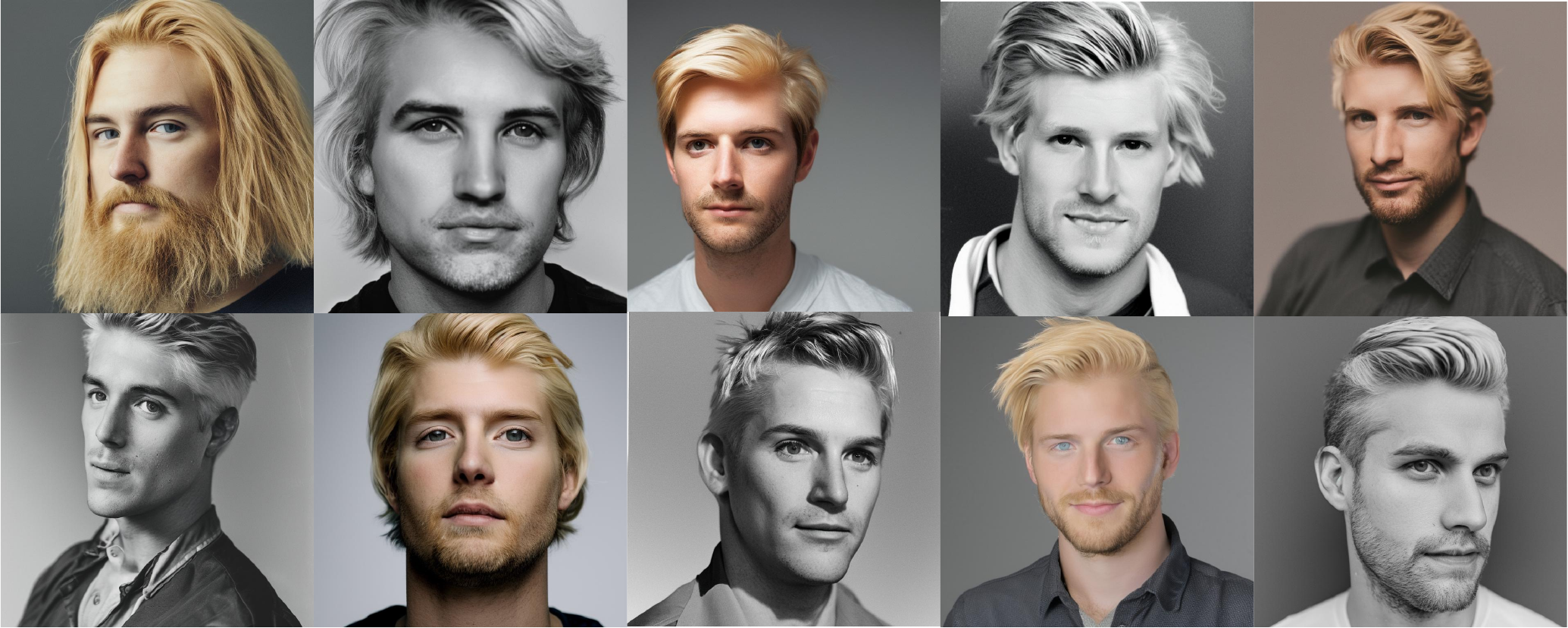}
  \caption{Additional visualization of generated images on CelebA dataset. We generate images of blonde man as the minority group.}
\label{fig:celeba_generated}
\end{figure}

\end{document}